\documentclass[conference]{IEEEtran}
\IEEEoverridecommandlockouts
\usepackage{cite}
\usepackage{amsmath,amssymb,amsfonts}
\usepackage[noend]{algpseudocode}
\usepackage[ruled,vlined,linesnumbered]{algorithm2e}
\usepackage{algorithmicx}
\usepackage{graphicx}
\usepackage{textcomp}
\usepackage{xcolor}
\newtheorem{definition}{Definition}
\newtheorem{theorem}{Theorem}
\newcommand*{\Scale}[2][4]{\scalebox{#1}{$#2$}}%

\usepackage{mathtools}
\newcommand{\defeq}{\vcentcolon=}

\def\BibTeX{{\rm B\kern-.05em{\sc i\kern-.025em b}\kern-.08em
    T\kern-.1667em\lower.7ex\hbox{E}\kern-.125emX}}
\DeclareMathOperator*{\argminA}{arg\,min} 
\DeclareMathOperator*{\maxA}{max} 
\usepackage{subcaption}
\begin{document}

\title{Privacy-Preserving Debiasing using Data Augmentation and Machine Unlearning}

\author{\IEEEauthorblockN{Zhixin Pan, Emma Andrews, Laura Chang, and Prabhat Mishra}
\IEEEauthorblockA{Department of Computer \& Information Science \& Engineering \\
University of Florida, Gainesville, Florida, USA \\
\{panzhixin, e.andrews, laurachang, prabhat\}@ufl.edu}
}

\maketitle

\begin{abstract}
Data augmentation is widely used to mitigate data bias in the training dataset. However, data augmentation exposes machine learning models to privacy attacks, such as membership inference attacks. In this paper, we propose an effective combination of data augmentation and machine unlearning, which can reduce data bias while providing a provable defense against known attacks. 
Specifically, we maintain the fairness of the trained model with diffusion-based data augmentation, and then utilize multi-shard unlearning to remove identifying information of original data from the ML model for protection against privacy attacks. Experimental evaluation across diverse datasets demonstrates that our approach can achieve significant improvements in bias reduction as well as robustness against state-of-the-art privacy attacks.
\end{abstract}
\section{Introduction}
\label{sec:intro}
Deep Learning (DL) has emerged as a powerful tool for solving large-scale tasks across various applications. Its ability to learn intricate patterns and representations from data has made it a popular choice for many complex tasks. However, one of the challenges associated with DL models is the presence of inherent biases in training datasets. These biases can arise due to several factors, such as preferences of data collectors, societal biases, or imbalanced data distributions, among others. Biased training datasets can lead to unfair predictions and perpetuate discrimination when deployed in real-world applications.

Data augmentation techniques aim to reduce or eliminate biases present in the training data, thereby enabling more equitable and unbiased predictions. However, recent studies have highlighted the vulnerability of machine learning (ML) models trained on augmented datasets to membership inference attacks (MIAs)~\cite{yu2021does}. MIAs involve an adversary attempting to determine whether a particular data point was part of the training dataset used to train the model. Such attacks can compromise the privacy of individuals and undermine the security of ML systems.

\begin{figure}[htp]
    \centering
    \includegraphics[width = 0.45\textwidth]{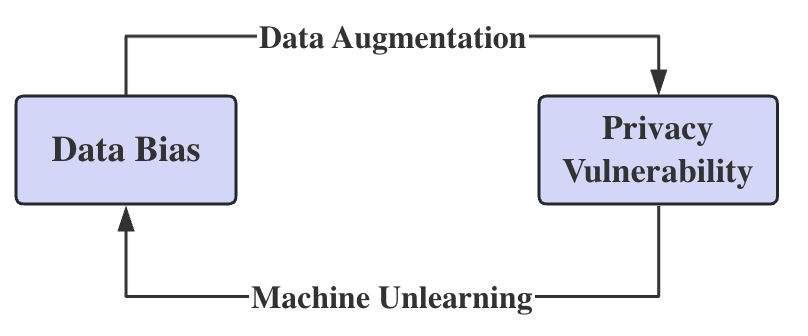}
    \vspace{-0.1in}
    \caption{Integration of data augmentation and machine unlearning can reduce both data bias and privacy attacks.}
    \label{fig:mutual}
    \vspace{-0.15 in}
\end{figure}

In this paper, we explore a synergistic integration of data augmentation and machine unlearning to address both fairness and privacy concerns as shown in Figure~\ref{fig:mutual}. Note that a naive combination of data augmentation and machine unlearning will not meet both privacy and fairness objectives. In fact, they can negatively influence each other (with augmentation aiding MIA, and unlearning increasing bias) as outlined in proposed method section. Specifically, this paper makes three important contributions.
\begin{itemize}
\item To the best of our knowledge, our approach is the first attempt at integrating data augmentation and machine unlearning algorithms. 
\item We show that the ML model produced by our proposed step-wise unlearning algorithm, synchronized with diffusion-based data augmentation, can achieve both fairness (bias reduction) and robustness (privacy-preservation) at the same time. 

\item Extensive experimental evaluation on diverse datasets demonstrates that our framework can significantly reduce data bias and improve the robustness of ML models against state-of-the-art MIA attacks.
\end{itemize}


\section{Background and Preliminaries}
\label{sec:relwork}
This section describes related efforts in data bias as well as membership inference attacks. 
\subsection{Data Bias and Countermeasures}
\begin{figure}[htp]
    \centering
    \includegraphics[width = 0.5\textwidth]{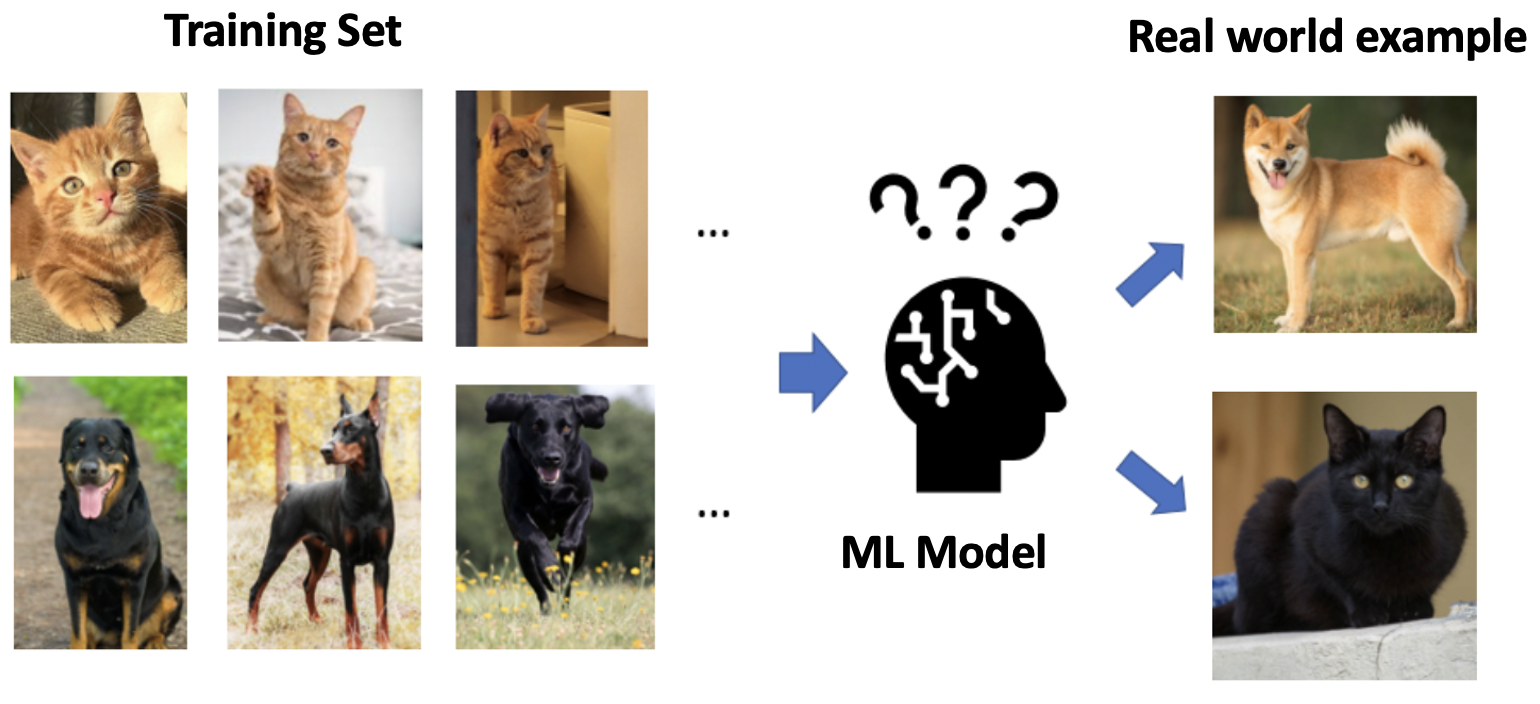}
    \vspace{-0.3in}
    \caption{An ML model trained with a biased feature distribution (e.g., no black cat) can lead to mispredictions.}
    \label{fig:biasExample}
\end{figure}

Data bias pertains to the existence of inequitable or discriminatory outcomes observed in the predictions or decisions made by machine learning models, caused by biased training data. This problem arises when the training dataset employed to construct the model inadequately represents the broader real-world population or inadvertently embodies historical biases and prejudices. Figure~\ref{fig:biasExample} illustrates a model's misprediction resulting from training data bias. In this example, only black dogs and orange cats are fed during the training phase. As a result, when black cats and orange dogs were given, the trained model is likely to make wrong decisions.

Existing bias reduction techniques can be divided into two main categories. The first category explores algorithmic improvements to explicitly address bias in machine learning. For example, a fairness-aware learning algorithm that aims to minimize bias in the decision-making process is proposed in ~\cite{kamishima2011fairness} by considering fairness as an explicit objective during model training. A fairness-ensured feature selection algorithm is proposed in~\cite{khodadadian2021information}. The second category performs data pre-processing, including data augmentation, oversampling, and under-sampling. The goal is to produce synthetic data samples to balance the representation of different groups in the training data. SMOTE \cite{chawla2002smote} uses over-sampling to even out the occurrences of each class. For data augmentation, generative model-based algorithms are widely applied with promising debiasing results were achieved \cite{amini2019uncovering,frid2018synthetic,mariani2018bagan}. However, existing data augmentation techniques usually rely on human-expert knowledge. Thus, manual crafting is widely applied, however, it lacks the guidance and measure of the bias extent. Also, a recent study~\cite{yu2021does} revealed that data augmentation inherently exposes ML models to membership inference attacks.

\subsection{Membership Inference Attack}
A membership inference (MIA) attack is a privacy attack that aims to determine whether a specific data sample was part of the training dataset used to build an ML model. The goal of such an attack is to exploit the model's behavior to infer membership information, thereby violating the privacy of individuals whose data was used for training.
%
%
There are various efforts to defend against MI attacks, including differential privacy (DP)~\cite{dwork2006differential}, regularization~\cite{kaya2020effectiveness}, and adversarial training~\cite{song2019membership}. However, a recent study~\cite{hu2022membership} has shown that none of the above solutions can completely eliminate the risk of membership inference attacks. While implementing a combination of these countermeasures might help, it is expensive in terms of time and memory. Also, in~\cite{yu2021does}, the author proposed a novel MIA which exploits the vulnerability of data augmentation to bypass the state-of-the-art detection approaches. The augmentation technique applied in~\cite{yu2021does} is image cropping, but the fundamental principle is identical with augmentation using diffusion models: \textit{data augmentation leverages information from the original dataset to create augmented samples}. This process inherently exposes more information from the original data, potentially making it more vulnerable to membership inference attacks. Our proposed method adopts the idea of machine unlearning to address this challenge.

\subsection{Notations and Preliminaries}
The primary objective of machine learning algorithms is to figure out optimal model parameters to embody a mapping from instance space to target space. Formally, let $\mathcal{D}$ be a distribution over the instance space $Z$, and $\mathcal{W} \subseteq \mathbb{R}^n$ be the parameter space. Consider $l: \mathcal{W} \times \mathcal{Z} \rightarrow \mathbb{R}$ be the loss function. The goal is to minimize the test loss population risk (test loss), which is given by 
\[
F(w) \defeq \mathbb{E}_{z\sim \mathcal{D}}[l(w,z)] 
\]
Solving the following optimization problem
\[F^* = \argminA_{w\in \mathcal{W}} F(w)\] 
gives the value of minimum loss and $w^*$ is the corresponding minimizer. In a realistic scenario, the distribution $\mathcal{D}$ is unknown and we usually apply a subset $S$ of i.i.d samples to find the approximation. 


\section{Proposed Method}
\label{sec:proposed}

In this section, we first discuss why a naive combination of data augmentation and machine unlearning is not sufficient to reduce data bias as well as defend against privacy attacks. 
The fundamental problem with a naive combination is that data augmentation and machine unlearning are not independent operations. In fact, they negatively influence each other. We briefly describe these scenarios to highlight the contributions of our proposed framework.  

\textbf{Augmentation Affects Privacy:} It is a common belief that data augmentation can protect against MIA since the existence of similar instances in the training set increase the difficulty of inferring the exact information of a target member. However, according to a recent study in~\cite{yu2021does}, the authors demonstrated 20\% better attack success rate when the model is trained with data augmentation. Formally, let $\mathcal{T}$ be the set of all possible transformations from original data sample to augmented samples. For a given data point $z$, each transformation $t \in \mathcal{T}$ generates an augmented instance $\tilde{z} = t(z)$. The size of $\mathcal{T}$ is usually infinite and a subset $T \subset \mathcal{T}$ is applied in practice. Then the learning objective with data augmentation is 
\[
w^* = \argminA_{w} (\sum\limits_{z \in \mathcal{Z}} \sum\limits_{\tilde{z} \in t(\mathcal{Z})} l(w, \tilde{z}) + \sum\limits_{z \in \mathcal{Z}} l(w, z))
\]
Suppose the size of dataset is $n$, then the MIA can be defined as deciding membership using $n$ i.i.d. Bernoulli samples $\{m_1,...,m_n\}$
When augmentation is applied, the authors in~\cite{yu2021does} have proved that the process 
\[
\{z_i\} \rightarrow \{t(z_i)\} \rightarrow \{w, m_i\}
\]
is a \textit{Markov chain} and hence by the conditional entropy theorem in~\cite{ash2012information}, the following holds:
\[
H(m_i| w,T(z_i)) = H(m_i|w,T(zi),di) \geq H(m_i| w,z_i)
\]
where $H(\cdot|\cdot)$ is the conditional entropy. The first equality is due to the Markov chain and the second inequality is due to the property of conditional entropy. This indicates that less uncertainty of $m_i$ can be obtained by $\{w, T(z_i)\}$ than by $\{w,z_i\}$, i.e, augmented dataset can inherently expose more information about the real data instances to the attacker.

\textbf{Unlearning Affects Bias:} Machine unlearning can inherently increase the possibility of making biased decisions by the retrained model. It could negatively influence debiasing by leaving information in the model inadvertently that would affect the bias metric being used \cite{chen2021when}. For example, with the class imbalance metric, if information on a specific class label is retained after unlearning, the debiasing strategy may not be able to detect this leftover bias and remove it effectively. Essentially, this problem is caused by that privacy-preserving machine unlearning algorithm are most defined based on $\epsilon-\delta$ differential privacy (DP) concept~\cite{dwork2006differential}, in a way that:
\begin{definition}
($\epsilon-\delta$ unlearning). For all set $S$ of size $n$ and delete requests $U \subseteq S$ such that $|U| \leq m$, and $W \subseteq \mathcal{W}$, a learning algorithm $A$ and an unlearning algorithm $\bar{A}$ is $\epsilon-\delta$-unlearning if:
\[\Scale[0.9]{
Pr(\bar{A}(U,A(S)) \in W \leq e^\epsilon \cdot Pr\bar{A}(\emptyset,A(S\backslash U)) \in W ) + \delta}
\]
\end{definition}

The main purpose of DP-based machine unlearning is to create a scenario where, with high probability, an observer cannot distinguish between two scenarios: (i) the model trained on the set $S$ and then $m$ points are removed using the unlearning algorithm with statistics $T(S)$, and (ii) the model trained on the set $S\backslash U$ without any subsequent point removal by the unlearning algorithm. 
However, if the remaining dataset after unlearning is biased, the retrained model is at a higher risk of making biased decisions. This scenario also creates an opportunity for potential attackers to intentionally remove specific records and inject stealthy bias backdoor into the model. Additionally, it has been proven that an unlearning algorithm can only unlearn up to $O(n/d^{1/4})$ samples~\cite{sekhari2021remember}, where $d$ is the problem dimension, i.e., existing unlearning algorithm cannot remove all the original data points.


 

\begin{figure}[htp]
\centering
\vspace{-0.1in}
\includegraphics[width = 0.52\textwidth]{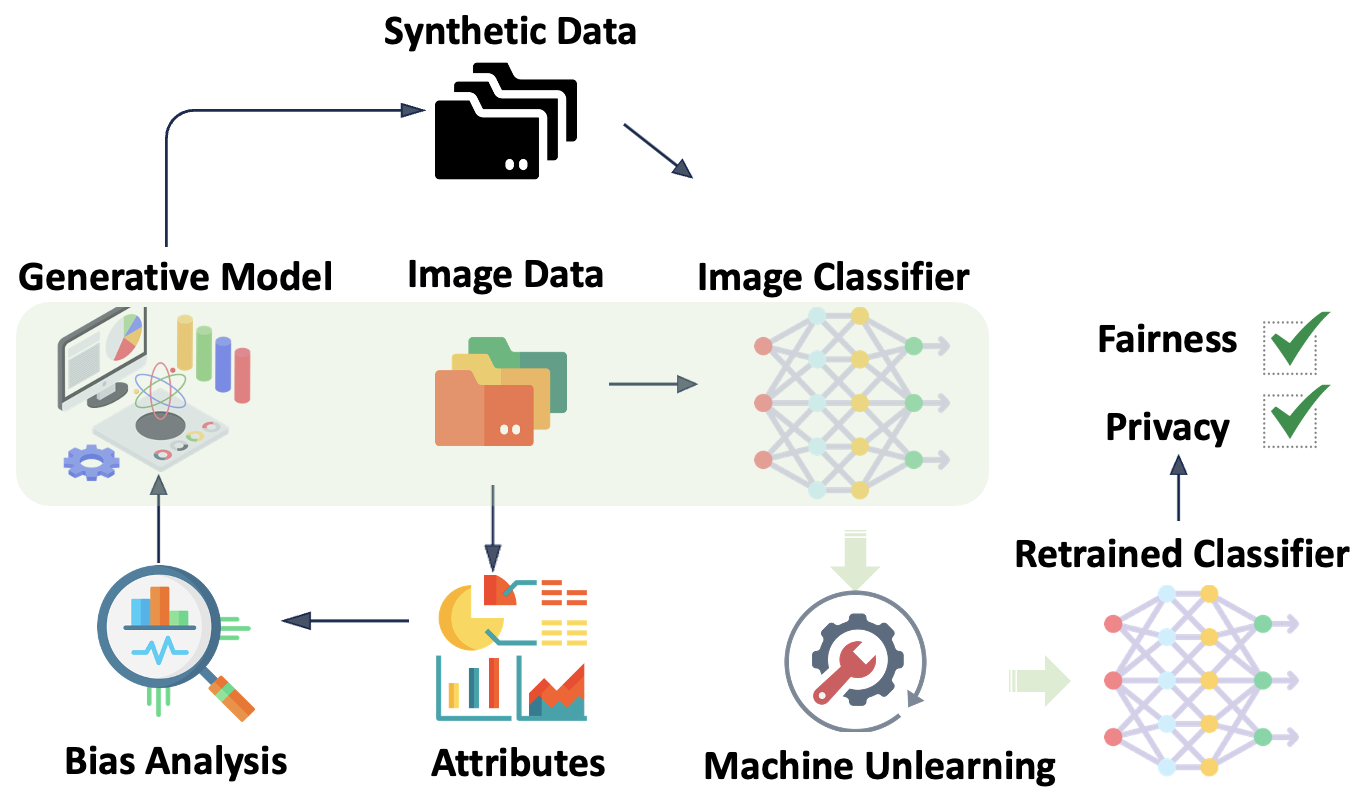}
\vspace{-0.1in}
\vspace{-0.1in}
\caption{Proposed framework with data augmentation and machine unlearning can achieve both privacy and fairness.}
\label{fig:figcmp}
\vspace{-0.1in}
\end{figure}

\subsection{Overview}
\label{sec:overview}
To address the above challenges, our proposed approach enables a synergistic integration of data augmentation and machine unlearning to achieve a privacy-preserving de-biasing framework. Figure~\ref{fig:figcmp} provides an overview of our proposed framework that consists of two major tasks. The first task performs \textit{data augmentation} performs guided diffusion based on a comprehensive bias measurement. The second task performs \textit{machine unlearning} that retrains the task classifier to partially forget original data points. The retrained classifier is developed in a distributed manner to achieve better time efficiency. We perform both tasks in each iteration, and repeat until convergence. Experimental results in the next section demonstrate that our iterative synchronization framework is capable of ensuring both privacy and fairness of the ML model. 





\subsection{Guided Diffusion Model based Augmentation}
\label{sec:dataaug}

Data augmentation can be performed by either manually crafting, or through generative models to learn the underlying distribution and generate new synthetic samples. Manual crafting is usually limited by human knowledge, and might not cover all necessary variations and patterns in the data. Therefore, we apply generative model in proposed framework. Specifically, we apply \textit{guided-diffusion model}.

For a data distribution $p_0(z_0)$ where $x_0 \in \mathcal{Z}$, a family of distributions $p_t(z_t)$ can be defined by injecting i.i.d. Gaussian noise to data samples, such that $z_t = z_0 + \sigma_t \eta$ with $\eta \sim \mathcal{N}(0,{\bf I})$, and $\sigma_t$ monotonically increasing with time $t \in [0,T]$. The score function w.r.t $z_0$ can be learned via a de-noising score matching objective: $L^0_t(\theta) = \mathbb{E}_{z_0,z_t} [||D_\theta(z_t,t)-z_0||_2^2]$. 
The guided diffusion model is an extension of the traditional diffusion model that allows for more control over the generated samples. It enables the generation of samples with specific desired properties. To apply guided diffusion in de-biasing task, the key problem is how do we accurately detect the cause and extend of bias so that we have a clear guideline for synthesizing data?



To answer the question, we need a careful design of bias metric. Let us conduct an in-depth analysis of the example in Figure~\ref{fig:biasExample}. 
Since the ML model does not understand the concept of \textit{cat} and \textit{dog}. It relies on knowledge gained from the training dataset, where the labels \textit{cat} and \textit{dog} simply represent the labels of two distinct categories. In other words, the ML model was trained to differentiate between these object categories, and owing to the biased distribution of colors among the classes, its  objective has been inherently transformed into distinguishing black and orange objects. Consequently, what we identify as a ``misprediction" in this instance is, in fact, a ``mismatch" in the understanding of tasks between the model and the expectations of users.
\color{black}

In our example, this mismatch was observed when the test phase presented a black cat or orange dog, which exists in the real world but was never encountered in the training set. This situation raises a controversial point: what if, in reality, all cats were actually orange and dogs were black? Under such an assumption, does the model still exhibit flaws? Addressing these questions in turn provides us with a comprehensive understanding of training data bias. 
Fundamentally, {the truth of training data bias is the distinction in attribute or feature distribution between the available dataset and the prior knowledge}. Obtaining a suitable prior distribution is commonly a manageable task, as it can be estimated through statistical analysis or leveraging existing knowledge about the problem domain. For example, when dealing with gender distribution, a simple and widely accepted prior distribution assumption of 50-50 male-female ratio can be applied. Then the key problem is transformed as measuring the posteriors distribution and measures the distance between it with the prior. In this paper, we consider the KL-divergence as the metric.

\begin{definition}
Let $D(\tau)$ be the prior distribution of an attribute $\tau$ of all instances, and $P(\tau)$ be the posterior distribution of that from the training dataset $S$. We define the partial bias of $S$ w.r.t ${\tau}$ be: \[\frac{\partial B(S,D)}{\partial {\tau}}  = KL(D(\tau)||P(\tau))\], i.e., the KL-divergence between these two distributions.
\end{definition}
The integral of all partial bias loss can be incorporated into the total training loss so that it serves as the guidance for sample generation. 
Formally, let $G$ be the augmented dataset generated from the guided diffusion model, guided by attributes distribution $\Lambda$. Then the total loss $\hat{L}$ in one iteration of guided sample generation is
\[
\hat{L} \defeq \frac{1}{2} \int \Big\|\frac{\partial B(S \cup G(\Lambda),D)}{\partial \tau}\Big\|^2 d\tau
\]
In real applications, we usually only consider a limited number of attributes. To simply, assume for each instance we only consider $j$ different attributes, in this case $ \Lambda= \{P({\tau_1}),..., P({\tau_j})\}$, and the above definition can be simplified as
\[
\hat{L} \defeq   \frac{1}{2} \sum\limits^j_{i=1} ||\, KL(D(\tau_i)||P(\tau_i))\,||^2
\]

As a sanity check, if prior distribution $D$ is already known. Then plug in $\Lambda = D$ gives an immediate minimizer of $\hat{L}$. 
In realistic applications, sometimes the prior distribution $D$ is a well-known fact or can be derived from experience. An example would be the distribution of human heights. Under this circumstance, we have clear guidance for the diffusion model. For example, to balance the bias of height values, $\Lambda$ can be simply set as the Gaussian. Sometimes the exact form of $D$ cannot be derived. In our framework, a variety of Estimation of distribution algorithms (EDAs) including Univariate marginal distribution algorithm~\cite{pelikan1999bivariate} (UMDA), Mutual information maximizing input clustering (MIMIC)~\cite{muhlenbein1997equation}, and Compact genetic algorithm (cGA)~\cite{harik1999compact} are applied to compute a decent approximation for a given task. An overview of data augmentation is shown in Algorithm~\ref{alg:alg1}.

\begin{algorithm}[h]
\SetKwInOut{Input}{Input}
\SetKwInOut{Output}{Output}
\Input{Dataset ($S$), threshold ($\rho$), attributes ($\{\tau_j\}$), diffusion model ($G$), gradient stepsize ($\lambda$)}
\Output{Augmented dataset $S^*$} 
{\bf Initialize} $ \Lambda= \{P({\tau_1}),..., P({\tau_j})\}$\\
$D$ = $EDA(S)$\\
  \Repeat{$B(S,D) < \rho$}{
  $\hat{L} = \frac{1}{2} \sum\limits^j_{i=1} ||\, KL(D(\tau_i)||P(\tau_i))\,||^2$\\
  $S = S \cup G(\Lambda)$\\
  $\Theta(\Lambda') = \Theta(\Lambda) + \lambda \cdot \nabla \hat{L}$\\
   }
{\bf Return} $S^* = S$   
\caption{Augmentation using guided-diffusion.}
\label{alg:alg1}
\end{algorithm}

Notice that our applied diffusion model extracts and follows the feature distribution from the training data, and thus generated results closely adhere to the original characteristics of the underlying data. This means that the process does not introduce new bias.

\subsection{Machine Unlearning}
\label{sec:unlearning}
\begin{figure}
    \centering
    \includegraphics[width = 0.35\textwidth]{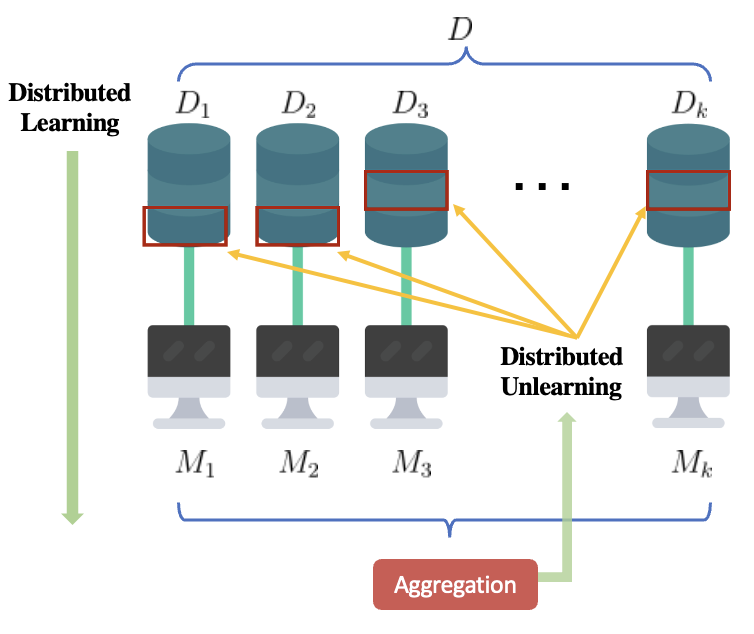}
        \vspace{-0.1 in}
    \caption{Our ML framework can compute both learning and unlearning in a parallel and distributed fashion.}
    \label{fig:distributed}
    \vspace{-0.1 in}
\end{figure}

We utilize machine unlearning for removing the learned knowledge or original data from the target ML model. Intuitively, by removing all the data records of original training dataset, we completely block all the possibility of MIA exposing private data. However, the naive way of  applying any existing machine unlearning approach suffers from two major problems: expensive resource cost, and inability of forgetting all data records.  
The first challenge of expensive cost is caused by the fact that unlearning algorithms typically require retraining the model from scratch, which can indeed be computationally expensive and time-consuming, especially for large-scale deep learning models. To address this challenge, we adopt the idea from~\cite{bourtoule2021machine} to perform unlearning in a distributed way, shown in Figure~\ref{fig:distributed}. Since both learning and unlearning are performed in a distributed manner, we can significantly speed up the unlearning process through parallel processing and distributed computing. 

The second shortcoming of directly applying machine unlearning is the limitation of deletion capacity. Intuitively, by deleting all records of original data completely mitigates MIA. However, there is a non-trivial limit on the deletion capacity an machine unlearning algorithm can obtain. Formally, the definition of deletion capacity is defined below:
\begin{definition}
Let $\epsilon, \delta \geq 0$. Let $S$ be a dataset of size $n$ drawn i.i.d. from $D$, and let $l(w,z)$ be a loss function. For a pair of learning and unlearning algorithms $A, \hat{A}$ that are $\epsilon,\delta$-unlearning, the deletion capacity $m_{\epsilon, \delta}^{A,\hat{A}}(d,n)$ is defined as the maximum number of samples $U$ that can be unlearnt, while still ensuring an excess population risk of 0.01.Specifically,
\[
m_{\epsilon, \delta}^{A,\hat{A}}(d,n) \defeq max \Bigl\{m| \mathbb{E} [\maxA F(\hat{A}(U,A(S))) - F^* ] \leq 0.01 \Bigl\}
\]
where $U \subset S\,\,$ and $ |U| \leq m$. 
\end{definition}

Based on the definition, a prior work in~\cite{sekhari2021remember} has proven the following theorem:
\begin{theorem}
Let $\epsilon = 1$ and $\delta < 0.005$. There exists a 4-Lipschitz, 1-strongly convex function $f$ and distribution $D$, such that for any learning algorithm $A$ and unlearning algorithm $\hat{A}$, the deletion capacity 
\[
m_{\epsilon, \delta}^{A,\hat{A}}(d,n)  \leq cn
\]
where $c$ depends on the properties of function $f$ and is strictly less than1.
\end{theorem}
The above result raises an immediate challenge to the privacy vulnerability, since there is an upper bound of deletion capacity without sacrificing the performance guarantee w.r.t. loss. 

We address this concern by apply data augmentation and machine unlearning alternatively in an recursive manner. The key idea is, at each iteration, we only unlearning $< cn$ samples to respect the tight upperbound. Also the unlearning process was immediately followed by a data augmentation to replenish newly generated data points. As a result, the size of dataset remains unchanged and each unlearning process strictly follows the deletion limitation. Note that the amount of original data points decreases monotonically, replaced by synthetic data points. This step-wise unlearning copes with data augmentation synchronously, and the entire process works in way that exactly matches the loop shown in Figure~\ref{fig:mutual}. Due to our distributed unlearning, for each model we only need to forget $< \frac{cn}{k}$ points, which further reduces the overhead. Algorithm~\ref{alg:alg2} shows the major steps of combining both data augmentation (Algorithm 1) and machine unlearning.

\begin{algorithm}[h]
\SetKwInOut{Input}{Input}
\SetKwInOut{Output}{Output}
\Input{Dataset ($S$), group number $k$, threshold $j$, $\epsilon-\delta$ learning-unlearning algorithms $A$ and $\hat{A}$}
\Output{Augmented dataset $S^*$, privacy-preserving classifier $M^*$} 
{\bf Initialize} $\lambda$,  $\rho$, $G$, and $\{\tau_i\}$, $iter = 1$\\
{\bf Group Split}: $S = \{S_1, S_2,..., S_k\}$\\ 
\Repeat{$iter \geq j$ or convergence}{
\For{$i$ from $1$ to $k$}{
$S'_i$ = {\bf Algorithm1}($S_i, \rho, \{\tau_i\}, G, \lambda$)\\
$M_i = A(S_i)$\\
{\bf Sample} $U_i \subset S_i$ and $ |U_i| \leq m_{\epsilon, \delta}^{A,\hat{A}}(|\{\tau_i\}|,|S_i|)$ \\
$M'_i = \hat{A}(S'_i, U_i)$\\
$S_i = S'_i - U_i$\\
$iter ++$
}
}
{\bf Return} $S^* = \{S_1, S_2,..., S_k\}$, $M^* = aggregate(M_1, M_2,..., M_k)$   
\caption{Iterative unlearning and augmentation.}
\label{alg:alg2}
\end{algorithm}

In this section, we present the experimental evaluation of our proposed framework. We will start with the introduction of experiment setup, then evaluate the performance of debasing as well as defending against MIA separately.


\section{Experimental Evaluation}
\label{sec:exp}
We first discuss experimental setup. Next, we present experimental results.
\subsection{Experimental Setup}

The data augmentation process using guided diffusion model was conducted on a system with an Intel Core i7-10700 processor, an NVIDIA 3070 Ti graphics card, and 16 GB of RAM. The operating system was Ubuntu 20.04, complemented by TensorFlow version 2.9.3 and Python 3.8.17. For the machine unlearning phase, we employed a workstation equipped with an AMD Ryzen 7 3700X processor, an NVIDIA RTX 2080 Super graphics card, and 64 GB of RAM.To facilitate model development and training, we leveraged PyTorch version 2.0.1 and Python 3.11.4. 
The experiments involved training a ResNet18 architecture on two distinct datasets, namely CIFAR-10 and CelebA. By leveraging transfer learning, we aimed to exploit the pre-trained features of the ResNet18 architecture, fine-tuning it to our specific classification objectives. To simplify the alternative training work, we set a forget rate of 10\% at each iteration, i.e., we forget 10\% of overall original data points in one turn, so that we implement the unlearning process in a step-wise manner while not violating the deletion capacity limitation outlined in Theorem 1.

To demonstrate the privacy-preserving property of our proposed method, three different MIA attacks were explored:
\begin{itemize}
    \item {\bf M-Loss}: The state-of-the-art black-box membership inference attack~\cite{song2019privacy}.
    \item {\bf M-Mean}: The state-of-the-art mean-value based membership inference attack proposed in~\cite{song2019privacy}.
    \item {\bf Moments}: The state-of-the-art membership inference attack targeting the property of augmented dataset~\cite{yu2021does}.
\end{itemize}

\subsection{Reduction of Data Bias }
In this section, we present the evaluation of the proposed guided-diffusion model-based data augmentation technique aimed at mitigating data bias. We use CIFAR-10 for illustration. The primary goal of proposed approach is to simultaneously address two key aspects: (1) the reduction of class-imbalance and (2) the preservation of data quality through augmented samples.

\subsubsection{Class-Imbalance Reduction:} 
\begin{figure}[htbp]
    \centering
    \includegraphics[width = 0.35\textwidth]{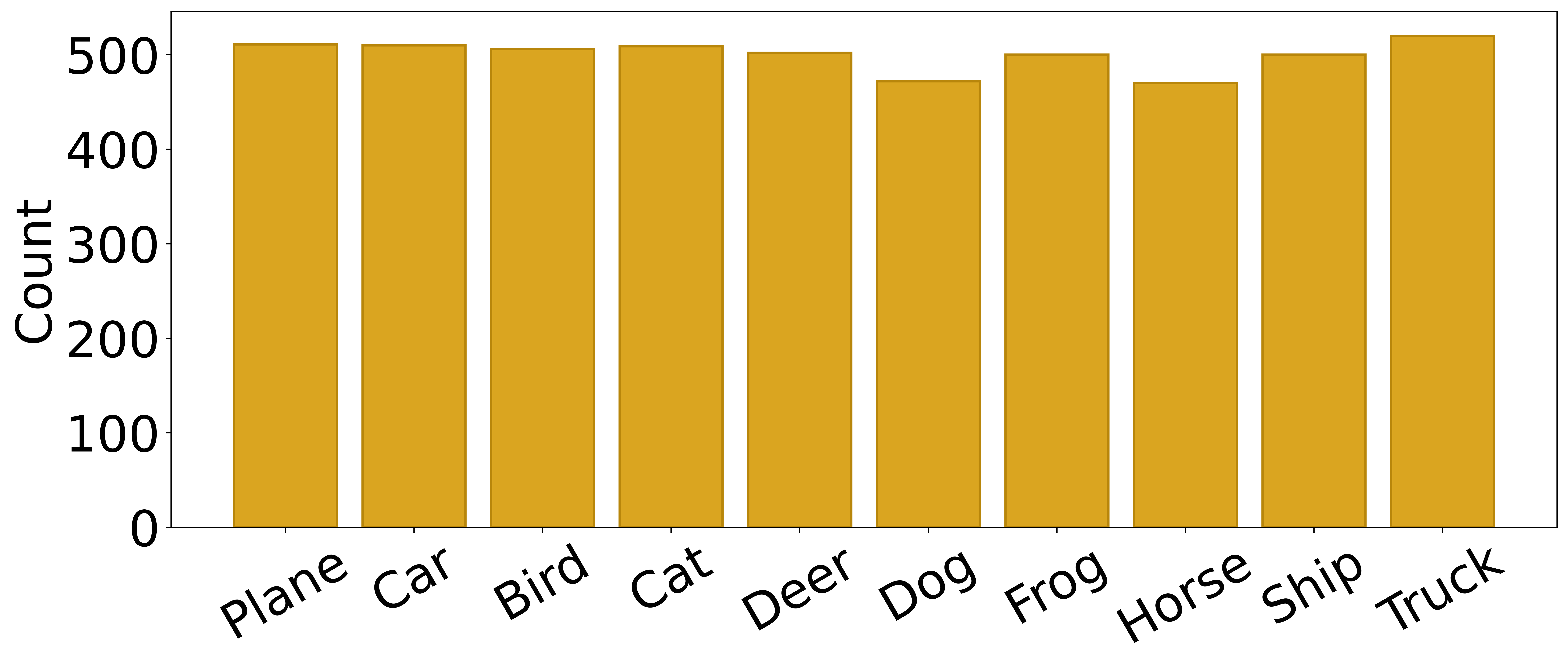}
    \caption{Distribution of individual classes in randomly selected subset of CIFAR-10 with 5000 samples}
    \label{fig:cifar10IndividualDist}
\end{figure}

Quantitative analysis reveals a significant reduction in class-imbalance across CIFAR-10 dataset. Intuitively, CIFAR-10 has an evenly distribution over 10 classes as shown in Figure~\ref{fig:cifar10IndividualDist}, which makes it seemly unbiased.

However, when we shift our focus to the distribution of superclass distribution (animals \& vehicles) instead of subcategories, the inherent class imbalance (animals:2959 vs vehicles: 2041) was observed.


To mitigate this bias, we apply diffusion model to generate synthetic samples and bring the overall dataset closer to a balanced distribution. Specifically, we generate 918 vehicle images which follow the original subclasss distribution as in the original dataset. This enhancement in class balance is instrumental in promoting fair and unbiased model training, ultimately leading to improved generalization performance for next sections.

\begin{figure}[htp]
    \centering
    \includegraphics[width = 0.35\textwidth]{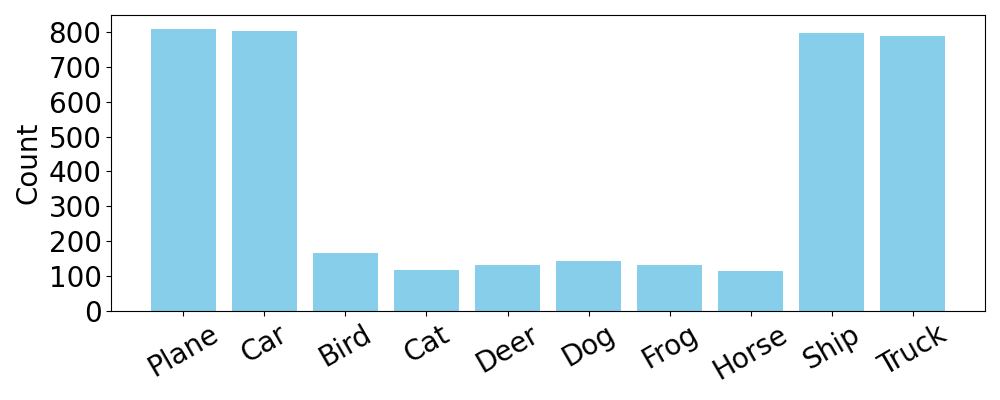}
    \caption{Biased training dataset distribution}
    \label{fig:trainingDistPreDebiasing1}
    \end{figure}

To illustrate the effect of proposed debiasing framework. We show the effects of debiasing by training a separate classifier before and after debiasing. Figure~\ref{fig:trainingDistPreDebiasing1} shows the bias before data augmentation (4000 training original cifar10 images with a 80/20 vehicle animal split, and 200 training iterations). 
    
\begin{figure}[htp]
    \centering
    \includegraphics[width = 0.35\textwidth]{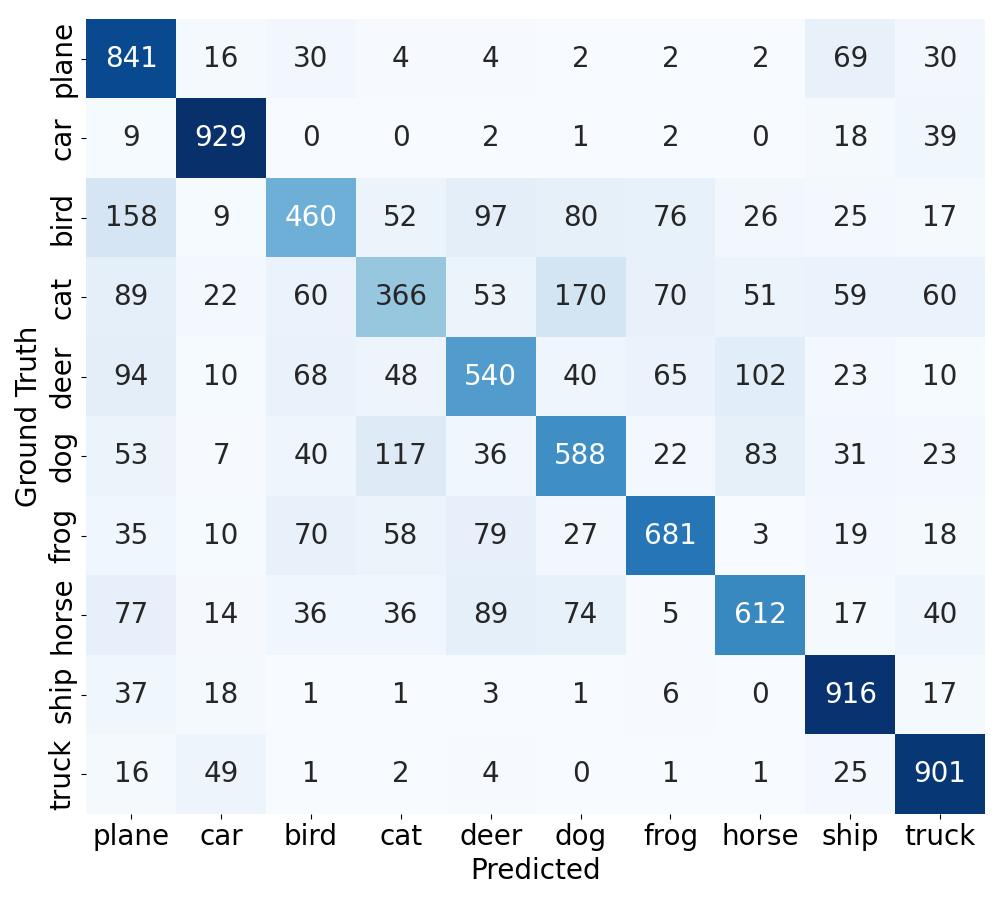}
    \caption{Classifier performance on biased dataset}
    \label{fig:classifierPreDebiasing1}
    \end{figure}

As shown in Figure~\ref{fig:classifierPreDebiasing1}, the classifier performs well for vehicles, however, there is a significant drop in performance for the animal categories. This is expected since the original dataset is heavily biased towards the vehicles.
    
\begin{figure}[htp]
    \centering
    \includegraphics[width = 0.35\textwidth]{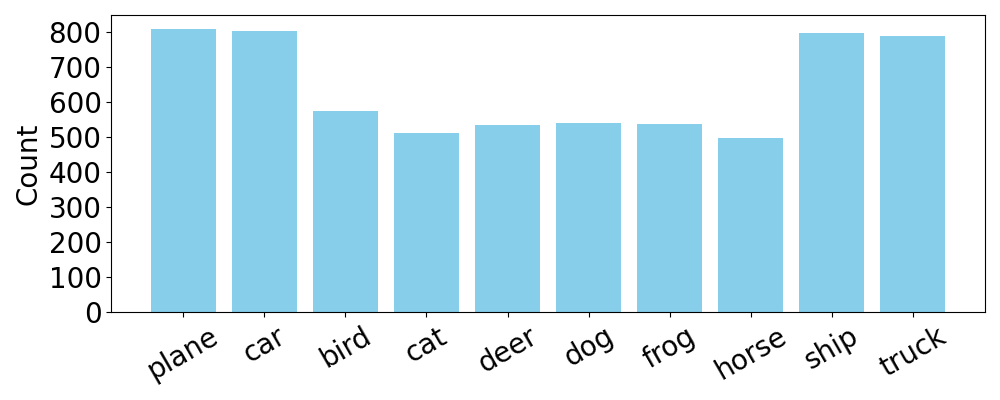}
    \caption{debiased training dataset distribution}
    \label{fig:trainingDistPostDebiasing1}
    \end{figure}

Now, we analyze the scenario after debiasing (2400 additional synthetic cifar10 images added to original,  50/50 vehicle animal split, 200 training iterations). Figure~\ref{fig:trainingDistPostDebiasing1} shows the dataset after applying our proposed data augmentation, which provides a better distribution compared to Figure~\ref{fig:trainingDistPreDebiasing1}.

\begin{figure}[htp]
    \centering
    \includegraphics[width = 0.35\textwidth]{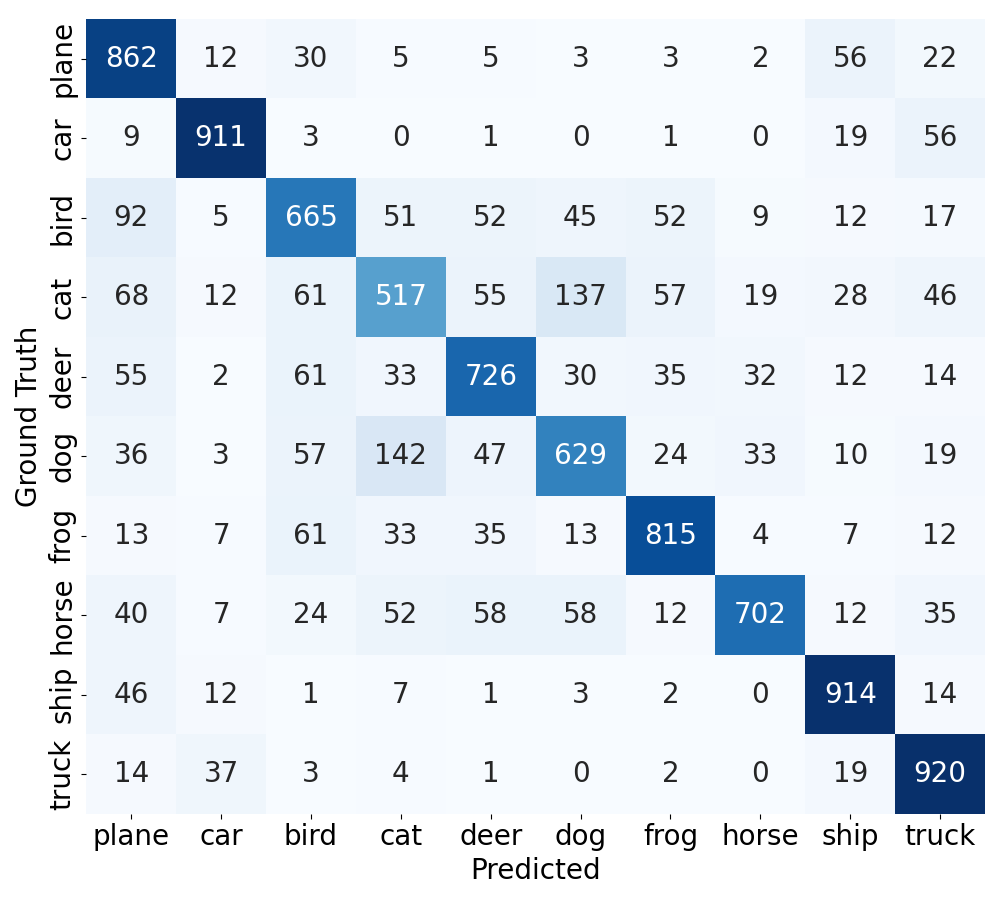}
    \caption{Classifier performance on debiased dataset}
    \label{fig:classifierPostDebiasing1}
    \end{figure}

As expected, the classifier performance shown in Figure~\ref{fig:classifierPostDebiasing1} is superior for each category compared to Figure~\ref{fig:classifierPreDebiasing1}.
    
\subsubsection{Preservation of Data Quality}

\begin{figure}[htp]
    \centering
    \includegraphics[width = 0.39\textwidth]{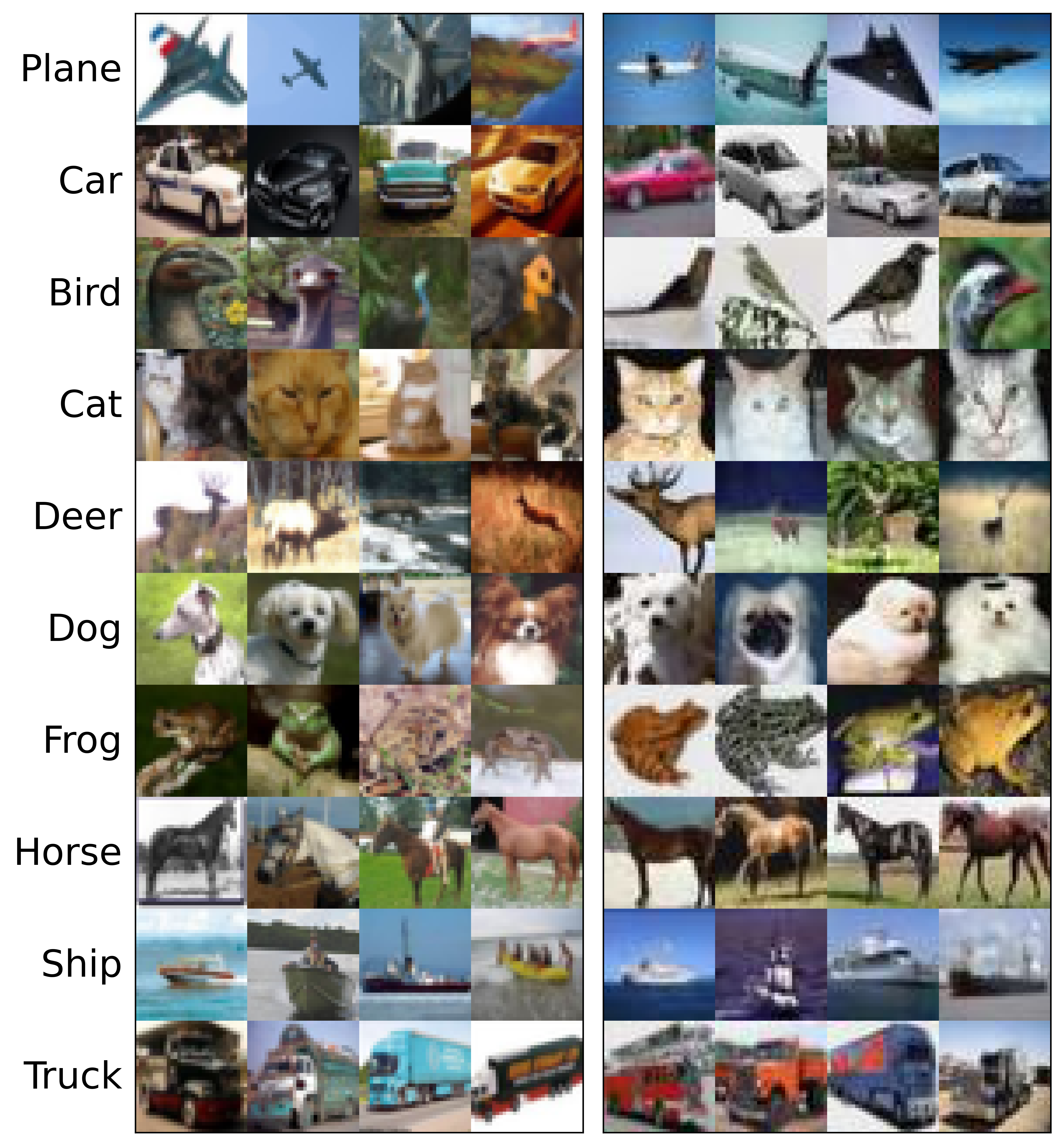}
    \caption{Batch of original (left) and synthetic (right) CIFAR-10 images. Each row represents a different class.}
    \label{fig:cifar10ImageQualityComparison}
\end{figure}

\begin{figure}[htp]
    \centering
    \includegraphics[width=0.23\textwidth]{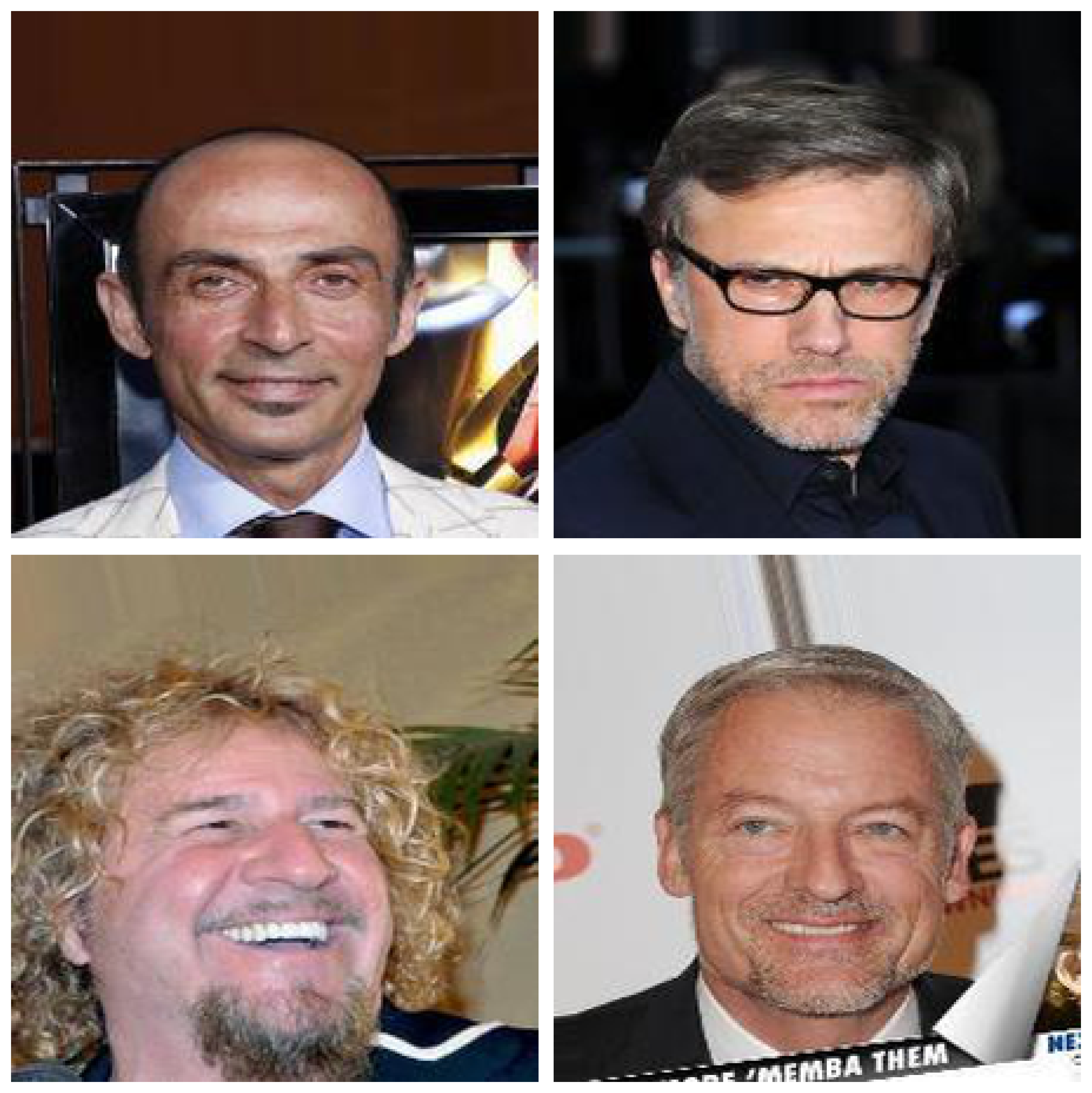}%
    \hfill
    \textcolor{black}{\vrule width 1pt}
    \includegraphics[width=0.23\textwidth]{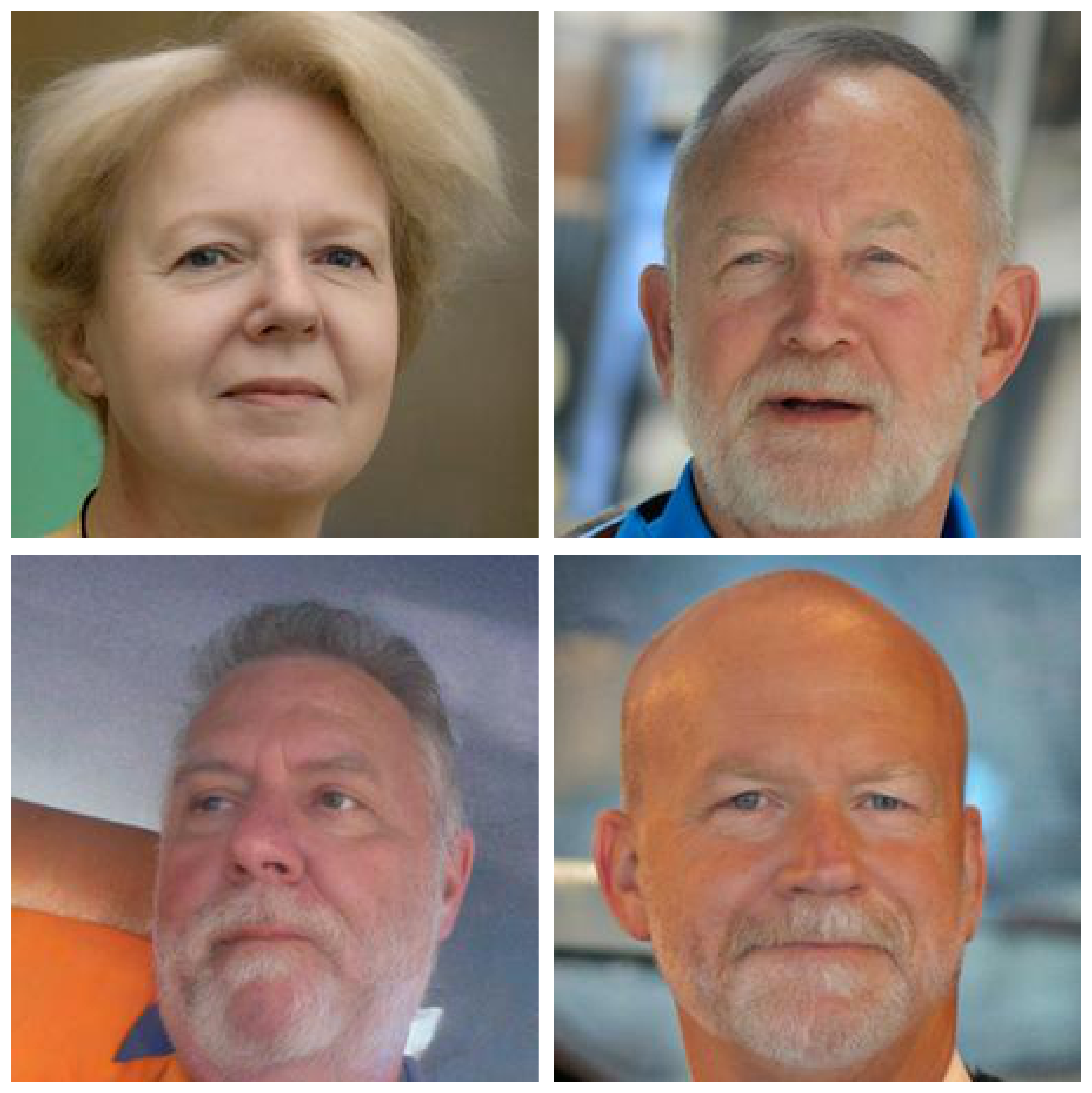}
    \caption{Batch of original (left) and synthetic (right) not 'Young' CelebA images}
    \label{fig:celebAImageQualityComparison}
\end{figure}

While addressing class-imbalance is crucial, maintaining the quality and integrity of augmented data samples is of paramount importance. In Figure~\ref{fig:cifar10ImageQualityComparison}, we compare one batch of original and generated CIFAR-10 images, where each row represents a specific class. Figure~\ref{fig:celebAImageQualityComparison} provides a similar display for CelebA images. The augmented samples have been guided against the `Young' attribute and presented for comparison against images from the original dataset labeled not 'Young'. As we can see, the augmented samples exhibit levels of quality on par with the original data.

\vspace{-0.2in}

\begin{table*}[htp]
\caption{The retain accuracy, testing accuracy, and corresponding training time of image classifier under different forget rate using alternative unlearning and data augmentation algorithm.}
\centering
\begin{tabular}{|c|c|c|c|c|c|c|c|c|c|c|}
\hline
\textbf{Iteration} & \textbf{1} & \textbf{2} & \textbf{3} & \textbf{4} & \textbf{5} & \textbf{6} & \textbf{7} & \textbf{8} & \textbf{9} & \textbf{10} \\ \hline
Retain  & 74.5\% & 74.4\% & 75.8\% & 75.6\% & 75.8\% & 74.6\% & 76.1\% & 74.4\% & 75.7\% & 74.1\% \\ \hline
Testing  & 64.6\% & 65.5\% & 64.2\% & 65.3\% & 65.3\% & 66.8\% & 65.0\% & 65.5\% & 68.8\% & 64.5\% \\ \hline
Training Time & 11m 1s & 12m 5s & 11m 48s & 11m 55s & 11m 9s & 10m 24s & 10m 29s & 10m 33s & 10m 54s & 11m 1s \\ \hline
\end{tabular}
\label{tab:retrain}
\end{table*}



\subsection{Performance of Model Unlearning}
Our step-wise unlearning framework demonstrates two critical attributes: (1) performance preservation, characterized by the maintenance of overall model accuracy despite data removal, and (2) privacy preservation, indicated by a substantial reduction in the success rate of state-of-the-art membership inference attacks.

\subsubsection{Performance Unchanged Unlearning} 
One of the key challenges in the domain of data privacy and security is the need to remove sensitive or outdated data from machine learning models without compromising their performance. In our experiment, we set a forget rate at 10\% for each alternative iteration. In Table \ref{tab:retrain}, we provide the training \& testing accuracy of models at each iteration. The results consistently show that our algorithm successfully maintains acceptable level of performance even after removing a subset of training samples, yet the model's accuracy on both training and validation sets remains stable. This highlights our proposed algorithm's ability to unlearn private data instances while minimally affecting model's predictive performance.

\subsection{Robustness against MIAs}
\label{sec:MIAs}
\begin{figure}[htp]
     \centering
     \begin{subfigure}[b]{0.35\columnwidth}
         \centering
         \includegraphics[width=\textwidth]{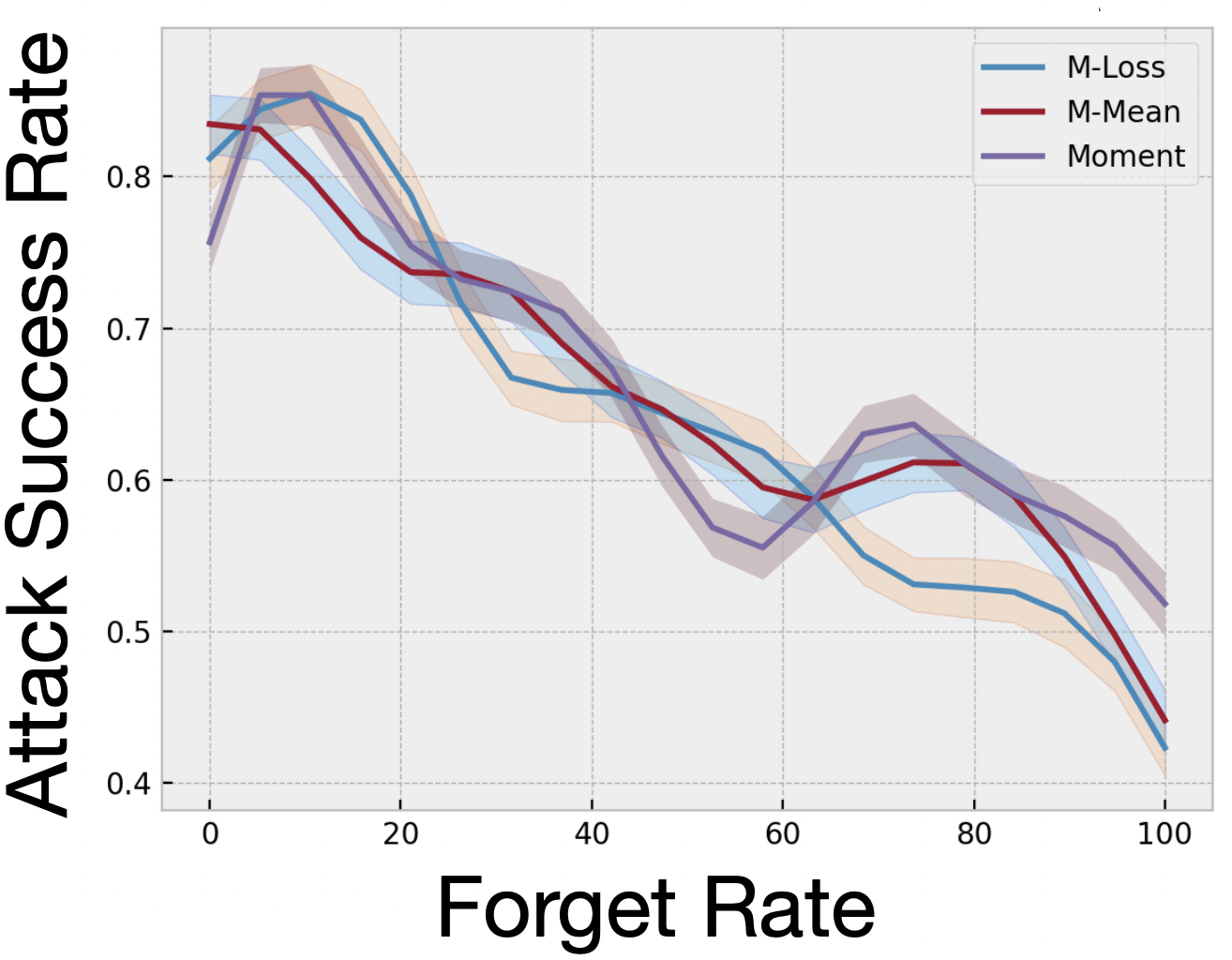}
         \caption{CIFAR-10}
         \label{fig:cifar}
     \end{subfigure}
     \hfill
     \begin{subfigure}[b]{0.35\columnwidth}
         \centering
         \includegraphics[width=\textwidth]{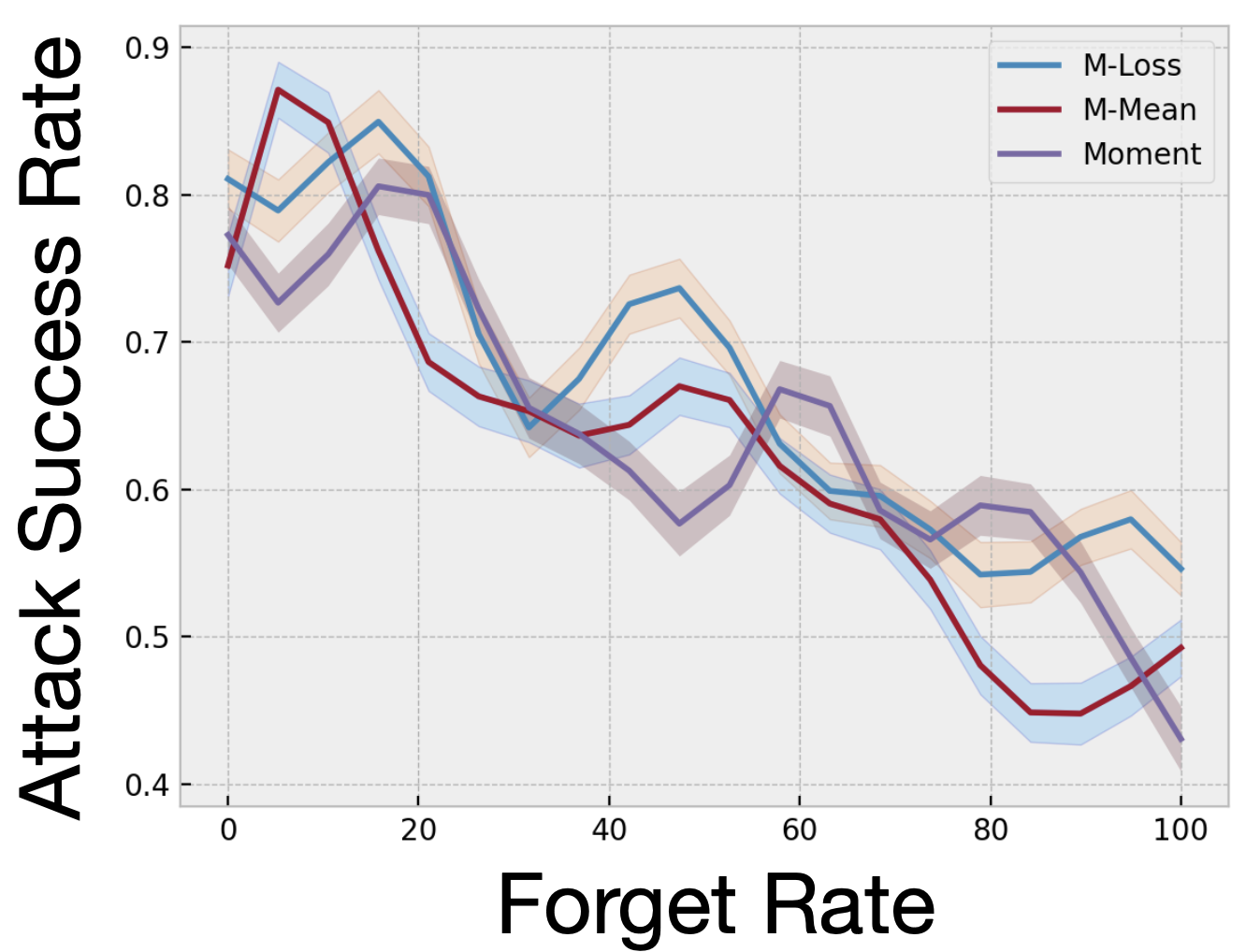}
         \caption{CelebA}
         \label{fig:celeb}
     \end{subfigure}
        \caption{MIA on CIFAR-10 and CelebA dataset under different forget rates.}
        \label{fig:mias}
\end{figure}

We rigorously evaluate the privacy-preserving capabilities of our algorithm by subjecting it to state-of-the-art membership inference attacks. 
In the context of the CIFAR-10 dataset, the baseline attack success rate directly applied on original dataset is 45\%. While from Figure~\ref{fig:mias}, we observe distinct trends in the order of success rates among the considered membership inference attacks. Specifically, the success rate hierarchy, in terms of Moment-based attack, M-Mean-based attack, and M-loss-based attack, is consistently observed to be Moment $>$ M-Mean $>$ M-loss. This phenomenon can be attributed to the intrinsic simplicity of the CIFAR-10 dataset, which lends itself to varying degrees of exploitability by the employed attacks. When evaluating the CelebA dataset, the baseline attack success rate holds as 37\%. While a divergent pattern emerges, whereby the sequence of success rates follows Moment $>$ M-loss $>$ M-mean. This shift in the attack success rate hierarchy may be attributed to the inherent intricacies introduced by the blackbox nature of the M-loss attack model. Especially, the Moment-based attack consistently maintains a success rate exceeding 60\%, even when subjected to an 80\% rate of data forgetting. However, this efficacy gradually diminishes to an approximate 50\% success rate over subsequent iterations, converging to a level akin to random guessing.

\section{Conclusion}
\label{sec:conclude}

In this paper, we presented a comprehensive framework that leverages the power of machine unlearning and data augmentation to address the critical issues of membership inference attacks (MIAs) and data bias in machine learning. By combining these two techniques, we have developed a privacy-preserving solution that enhances the robustness and reliability of modern machine learning models. We have theoretically analyzed the mutual impact between data bias and privacy of any given training dataset, and demonstrated that an iterative synchronization of these two techniques can simultaneously mitigate both data bias and privacy concerns. Specifically, the incorporation of data augmentation techniques mitigates the impact of data bias by artificially expanding the dataset through transformations and perturbations, while the utilization of machine unlearning prevents the risks associated with membership inference attacks by selectively removing individual data points from the training dataset. The effect of the synergistic integration has been evaluated through experiments, where we offer a robust ($<45\%$ attack success rate under MIA) ML solution in both CIFAR-10 and CelebA dataset, along with a balanced training dataset of high-quality synthetic samples using a guided diffusion model.

\bibliographystyle{IEEEtran}
\bibliography{IEEEabrv,refs.bib}

\end{document}